\documentclass[runningheads,a4paper]{styFiles/llncs}

\usepackage{amssymb}
\usepackage{graphicx}
\usepackage{amsmath}
\usepackage{styFiles/mydefs}
\usepackage{hyperref}
\usepackage{url}
\usepackage{breqn}
\usepackage{subfig}
\usepackage{url}

\usepackage{tikz}
\usetikzlibrary{decorations.pathmorphing} 
\usetikzlibrary{fit}					
\usetikzlibrary{backgrounds}	

\tikzstyle{every node}=[font=\Large]

\urldef{\mailsa}\path|{alfred.hofmann, ursula.barth, ingrid.haas, frank.holzwarth,|
\urldef{\mailsb}\path|anna.kramer, leonie.kunz, christine.reiss, nicole.sator,|
\urldef{\mailsc}\path|erika.siebert-cole, peter.strasser, lncs}@springer.com|    
\newcommand{\keywords}[1]{\par\addvspace\baselineskip
\noindent\keywordname\enspace\ignorespaces#1}

\begin{document}

\mainmatter  

\title{Multitask Learning for Sequence Labeling Tasks}

\titlerunning{Multitask Learning for Sequence Labeling Tasks}

%
%
\author{
Arvind Agarwal
\and Saurabh Kataria
}
\authorrunning{Multitask Learning for Sequence Labeling Tasks}

\institute{Xerox Research Center, Webster\\
Webster, New York 14580 USA\\
{\texttt \{arvind.agarwal,saurabh.kataria\}@exrox.com}\\
}

%
%

\toctitle{Lecture Notes in Computer Science}
\tocauthor{Authors' Instructions}
\maketitle

\begin{abstract}
In this paper, we present a learning method for sequence labeling tasks in which each example sequence has multiple label sequences.  Our method learns multiple models, one model for each label sequence. Each model computes the joint probability of all label sequences given the example sequence. Although each model considers all label sequences, its primary focus is only one label sequence, and therefore, each model becomes a task-specific model, for the task belonging to that primary label. Such multiple models are learned {\it simultaneously} by facilitating the learning transfer among models through {\it explicit parameter sharing}. We experiment the proposed method on two applications and show that our method significantly outperforms the state-of-the-art  method. 
\end{abstract}

\section{Introduction}
Sequence labeling is an important task that finds applications in many areas such as bio-informatics, natural language processing (NLP), speech recognition, image processing etc. Depending upon the underlying sequence labeling task, labels are assigned to the tokens present in the sequence. Often in many domains, multiple labeling tasks need to be specified for the same sequence, i.e., multiple task specific labels are assigned to each token in the same sequence. For example, in NLP domain, words in a sentence can be labeled with their Part of Speech (POS) tags as well as the phrase chunks \cite{tjong2000introduction}. In another domain, a customer-care center might be interested in labeling the textual conversations between a customer and customer-care agent with {\it resolution status of a product specific issue} as well as the semantic tone of conversations, a.k.a. {\it dialogue act} \cite{Stolcke:2000}. The resolution status of an issue may belong to one of many classes such as \texttt{OPEN}, \texttt{CLOSED} etc. while the dialogue act may belong to classes such as \texttt{COMPLAINT}, \texttt{REQUEST}, \texttt{ANSWER} etc. See \tabref{conv_example} and  \secref{exp} for further details. 

Multiple tasks formed on one sequence, typically, tend to have intrinsic inter-label correlations. For instance, in customer-care domain, customers typically have complaint in their tone while describing their issue with a certain product. Incidentally, issues status \texttt{open} tends to have correlation with dialogue class \texttt{COMPLAINT}.  Similarly, correlations may exist between dialogue act \texttt{ANSWER} and issue status \texttt{CLOSED}. An example of a conversation showing such correlations is given in \tabref{conv_example}.

\begin{table}[t]
\centering
\caption{{\small Example of a conversation between a customer and customer-care agent. Each conversation text (column 3) has two labels associated with it (column 1 and 2).}}
{\small \ttfamily
\begin{tabular}{|c|c|p{3in}|}
\hline 
{\bf Dialogue Act} & {\bf Issue Status} & {\bf Text} \\ 
\hline 
COMPLAINT & OPEN & this mobile is making my life hard   \\
REQEST &  OPEN & @user\_x is there anything we in the social media  team can assist you with ?   \\
COMPLAINT & OPEN & so i can't make or receive calls \& i can't send texts or receive them.  \\ 
REQUEST & OPEN &  @user\_x what is your zip code ?  \\ 
ANNOUNCEMENT & OPEN & we can check for outages for you. \\
ANSWER & SOLVED & @user\_x your experiencing tower outage. \\
ANSWER & SOLVED & it is estimated to be cleared by the 21st of february. \\ 
\hline 
\end{tabular} 
}
\label{tab:conv_example}
\end{table}

In such a setting where each token in a sequence has multiple labels and these multiple labels exhibit correlations, it is important that the learning algorithm takes advantage of these correlations. If we define a task as learning from pairs of example sequence and its corresponding label sequence, then we can cast learning multiple label sequences as multitask sequence labeling  learning problem. In machine learning, Multitask Learning(MTL) is a well known problem. It provides a mechanism to learn various related tasks simultaneously such that learning from one task can benefit other task and vice versa. Often in MTL, multiple tasks are learned together by sharing their parameters explicitly either in a Bayesian way \cite{liusmtl09,yugp05,xuedp07} or in a non-Bayesian way \cite{andreas07,Micchelli04regularizedmulti-task,jacob08}. In MTL, most of the work has focused on classification or regression problems, with very little work on sequence labeling problem. In addition, most of the MTL methods are not especially designed for {\it our} multitask setting, i.e., an example sequence has multiple label sequences. Any method especially designed for multiple label sequences setting should exploit the dependencies among labels. 
One recent work that exploits the label dependencies is the work of \cite{sutton2007dynamic}. In this work, authors build a model called factorial CRF (appropriate for sequence labeling tasks) that is a {\it combined} CRF model {\it implicitly} learned on multiple tasks. In contrast to this method, our proposed method exploits the correlations present in multiple label sequences {\it explicitly} that not only improves upon the factorial CRF but also leads to a flexible framework for multitask sequence learning.  

In this work, we extend the MTL setting to the sequence labeling problem with multiple label sequences, and propose a new method for learning from multiple sequence labeling tasks simultaneously. Our method --- based on conditional random fields (CRFs) --- not only exploits label dependencies but also learns multiple tasks simultaneously by {\it explicitly} sharing parameters. In our method, we learn one model for each task \footnote{A task definition is expanded to include all labels. A task, for our method, is defined as learning from tuples of example sequence and its label sequences. Each task has one  {\it primary} label sequence, and other label sequences are considered {\it secondary}.}. Each task has two factors (as opposed to one factor in CRFs), one factor corresponding to {\it all} labels ( we call it {\it label dependency factor}), and other factor corresponding to task-specific {\it primary} label (we call it {\it task-specific factor}). Since the factor corresponding to {\it all} labels appear in all tasks, we facilitate the learning transfer among tasks by keeping the parameters corresponding to this factor same across all tasks. We show through a variety of experiments on two different data sets that such a model outperforms the state-of-the-art method. Note that learning from multiple labels is typically done in two ways: (1) build one single model that incorporates factors of all label sequences and example sequence, i.e., complete dependency and no independent learning (2) build multiple CRF-like {\it independent} models with no learning transfer among models, i.e., complete independent learning, and no dependency among labels. Our proposed is a middle ground between these two extremes, and provides the best of both worlds. Because of a task-specific factor, it allows model to learn independently, and at the same time, because of label-dependency factor, it allows learning to transferred among all tasks. 

In this work, we also propose a variation of this method where label dependency factor is further broken into two parts, one that will have information specific to the task and other information common to all tasks. This variation allows one to control the amount of transfer among multiple tasks. Experimental results of this variation show further improvement on two real world datasets. 


Our main contributions in this work are as follows: (1) we propose a new method for sequence labeling problem when a sequence has multiple labels.  We show the application of our method on multiple real world problems and show significant improvements. (2) The proposed method is simple and require little modification to the existing CRF method. (3) We propose a variation of this model that adds flexibility in terms of allowing one to control the amount of transfer among tasks. (4) The proposed method is naturally applicable to semi-supervised setting. It provides multiple models that can be used in co-training to incorporate unlabeled examples.

\section{Background and Problem Description}
\begin{figure}[htb]
\centering
\resizebox{1.9in}{1.2in}{%
\input{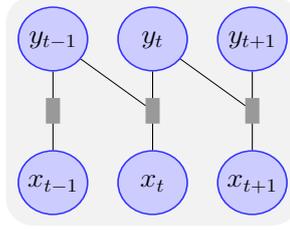}
}
\caption{\small Conditional Random Fields (CRFs)}
\label{fig:crf}
\end{figure}

We extend the mathematical framework of conditional random field (CRF) \cite{lafferty2001conditional} to support sequence labeling with multiple labels. Before describing our approach in detail, we first setup mathematical notations and summarize the CRF model. CRFs are undirected graphical models that model the conditional probability of a label sequence given an observed example sequence. Let $\mathcal{G}$ be an undirected graphical model over random variables $\x$ and $\y$ which represents sequence of random variables, i.e. $\x =(x_1,x_2,\dots x_T)$ is the sequence of observed entities (e.g. words in a sentence) that we want to label with $\y = (y_1,y_2,\dots y_T)$. $(\x,\y)$ together constitute an example-label pair. In the undirected graph $\mathcal{G}$, let $\mathcal{C} = \{C_1, C_2 \ldots \}$ be the set of cliques contained in the graph $\mathcal{G}$ where $C_i = \{\y_c,\x_c\}$, $\y_c \subset \y$ and $\x_c \subset \x$. Given such a graph defined on example-label pair, the conditional probability of a labeled sequence $\y$ given an observed example sequence $\x$ can be written as:
{\small
\begin{align}
p(\y|\x,\theta) = \frac{1}{Z(\x)} \prod_{c\in \mathcal{C}} \Phi(\y_c,\x_c|\theta),
\label{crf}
\end{align}
}
where $\Phi$ is the potential function defined over a clique, and is the function of all random variables in that clique. For example, in a specific case of linear chain CRF, these potential functions are defined over cliques $(x_t,y_{t-1},y_t)$. Here $\theta$ is the parameter, which we include in the potential function to denote that potential functions are parametric functions. $Z(\x) = \sum_{\y}\prod_{c\in \mathcal{C}} \Phi(\y_c,\x_c|\theta) $ is the partition function which makes sure that the potential functions are normalized, and \eqref{crf} can be interpreted as probabilities.  Usually the potential functions in \eqref{crf} factorize over the features of the clique and are defined using the exponential function of the form 
$\Phi(\y_c,\x_c|\theta) = \exp \left( \sum_k \theta_k f_k(\y_c,\x_c) \right) $. Here $f_k$ are the features functions, and $\theta_k$ are parameters. The feature functions $f_k$ can be defined arbitrarily which is one of the primary advantages of CRFs. For example, {\it part of speech} (POS) tagging problem can be modeled as  linear chain CRF, where feature functions can be defined over words, their characteristics, and their POS labels. In such a linear chain, indexed with $t$, a clique is defined for each $(y_{t-1},y_t,x_t)$) combination. For such a clique, one feature function could be a binary test: $f_k(y_{t - 1}, y_t, x_t)$ has value 1 if and only if $y_{t-1}$ has the label \textsc{ARTICLE}, $y_t$ has the label \textsc{NOUN}, and the word $x_t$ begins with a capital letter. A pictorial representation of CRF is given in \figref{crf}.

\subsection{Multitask Sequence Labeling}
In multitask sequence labeling problem, we are given multiple label sequences for each example sequence, i.e., in addition to $\y = (y_1,y_2,\dots y_T)$ (as defined for CRFs), we have $\z = (z_1,z_2,\dots z_T)$ as another set of label sequence for $\x$. For simplicity, we only consider two types of label sequences, however, it is straightforward to extend our approach to more than two labeling sequences (see \defnref{unshared}). Thus our training examples for the entire task become triplets of $(\x,\y,\z)$. We have $n$ such training examples, i.e., $\{\x_i,\y_i,\z_i\}_{i=1}^n$. Therefore, the multiple sequence labeling problem can be formalized as modeling conditional density $ p(\y,\z|\x)$.

\section{Our Approach}
\label{sec:approach}
In this section, we first describe a basic approach. When modeling the joint probability in multiple label setting, it is a standard practice to build just one model considering all possible factors (or cliques) from example sequence and label sequence \cite{sutton2007dynamic,marcheggiani2014hierarchical}. Although proven to be better that building two separate standard CRFs, this approach has many drawbacks (see experiments). One of them is the ability to model the tasks independently. The standard CRF though provides this capability, they do not include the effect of other labels; while other models, e.g., \cite{sutton2007dynamic,marcheggiani2014hierarchical} do not provide this capability at all -- they only build one single model. In our basic but novel approach, we begin by providing a middle ground between these two extremes (i.e. one single fully dependent model and two fully independent models), where both tasks are modeled independently but at the same time, one task draws benefit from other task through label dependencies. 

\begin{figure}[htb]
\resizebox {0.45\columnwidth} {!} {
\subfloat{
\tikzstyle{state}=[circle,
                                    thick,
                                    minimum size=1.2cm,
                                    draw=blue!80,
                                    fill=blue!20]

\tikzstyle{measurement}=[circle,
                                                thick,
                                                minimum size=1.2cm,
                                                draw=orange!80,
                                                fill=orange!25]

\tikzstyle{input}=[circle,
                                    thick,
                                    minimum size=1.2cm,
                                    draw=purple!80,
                                    fill=purple!20]

\tikzstyle{matrx}=[rectangle,
                                    thick,
                                    minimum size=0.1cm,
                                    draw=gray!80,
                                    fill=gray!80]

\tikzstyle{noise}=[circle,
                                    thick,
                                    minimum size=1.2cm,
                                    draw=yellow!85!black,
                                    fill=yellow!40,
                                    decorate,
                                    decoration={random steps,
                                                            segment length=2pt,
                                                            amplitude=2pt}]

\tikzstyle{background}=[rectangle,
                                                fill=gray!10,
                                                inner sep=0.2cm,
                                                rounded corners=5mm]

\begin{tikzpicture}[>=latex,text height=1.5ex,text depth=0.25ex]
  
  \matrix[row sep=0.5cm,column sep=0.5cm] {
    
        \node (1y_k-2) 		  {${}$};       &
        \node (1y_k-1) [state] {${z_{t-1}}$};       &
        \node (1em_k)   		  {${}$};     &
        \node (1y_k)   [state] {${z_{t}}$};       &
        \node (1em_k+1) 		  {${}$}; &
        \node (1y_k+1) [state] {${z_{t+1}}$};       &
        \\
        \node (1f_k-2) 		  {${}$}; &
        \node (1f_k-1) [matrx] {${}$}; &
        \node (1em_k) 		  {${}$};       &
        \node (1f_k)   [matrx] {${}$};     &
        \node (1em_k)   		  {${}$};       &
        \node (1f_k+1) [matrx] {${}$}; &
        \node (1f_k+2) 		  {${}$}; &
        \node (1f_k+3) 		  {${}$}; &
        \\
        \node (1x_k-2) 		  {${}$};       &
        \node (1x_k-1) [state] {${x_{t-1}}$};       &
        \node (1f1_k)   [matrx] {${}$};     &
        \node (1x_k)   [state] {${x_t}$};       &
        \node (1f1_k+1) [matrx] {${}$}; &
        \node (1x_k+1) [state] {${x_{t+1}}$};       &
        \node (1f1_k+2) [matrx] {${}$}; &
        \\
        
         \\
        \node (1z_k-2) 		  {${}$};       &
        \node (1z_k-1) [state] {${y_{t-1}}$};       &
        \node (1em_k)   		  {${}$};     &
        \node (1z_k)   [state] {${y_t}$};       &
        \node (1em_k+1) 	      {${}$}; &
        \node (1z_k+1) [state] {${y_{t+1}}$};       &
        \\

        \\        
        \node (z_k-2) 		  {${}$};       &
        \node (z_k-1) [state] {${z_{t-1}}$};       &
        \node (em_k)   		  {${}$};     &
        \node (z_k)   [state] {${z_{t}}$};       &
        \node (em_k+1) 		  {${}$}; &
        \node (z_k+1) [state] {${z_{t+1}}$};       &
        \\
        
        \\
        \node (x_k-2) 		  {${}$};       &
        \node (x_k-1) [state] {${x_{t-1}}$};       &
        \node (f1_k)   [matrx] {${}$};     &
        \node (x_k)   [state] {${x_t}$};       &
        \node (f1_k+1) [matrx] {${}$}; &
        \node (x_k+1) [state] {${x_{t+1}}$};       &
        \node (f1_k+2) [matrx] {${}$}; &
        \\
        
        \node (f_k-2) 		  {${}$}; &
        \node (f_k-1) [matrx] {${}$}; &
        \node (em_k) 		  {${}$};       &
        \node (f_k)   [matrx] {${}$};     &
        \node (em_k)   		  {${}$};       &
        \node (f_k+1) [matrx] {${}$}; &
        \node (f_k+2) 		  {${}$}; &
        \node (f_k+3) 		  {${}$}; &
        
         \\
        \node (y_k-2) 		  {${}$};       &
        \node (y_k-1) [state] {${y_{t-1}}$};       &
        \node (em_k)   		  {${}$};     &
        \node (y_k)   [state] {${y_t}$};       &
        \node (em_k+1) 	      {${}$}; &
        \node (y_k+1) [state] {${y_{t+1}}$};       &
        \\
    };
    
    \path[-]
        
		
		(1y_k-1) edge (1f1_k)	
		(1y_k-1) edge (1f_k-1)	
		(1y_k-1) edge (1f_k)
		(1x_k-1) edge (1f1_k)
		(1x_k-1) edge (1f_k-1)
		(1z_k-1) edge (1f1_k)

		(1y_k) edge (1f1_k+1)	
		(1y_k) edge (1f_k)	
		(1y_k) edge (1f_k+1)
		(1x_k) edge (1f1_k+1)
		(1x_k) edge (1f_k)
		(1z_k) edge (1f1_k+1)		
		
		(1y_k+1) edge (1f_k+1)	
		(1x_k+1) edge (1f_k+1)
		(1x_k+1) edge (1f1_k+2)
		(1y_k+1) edge (1f1_k+2)
		(1z_k+1) edge (1f1_k+2)	
		
		(1y_k+1) edge[dotted] (1f_k+3)	
		(1y_k-2) edge[dotted] (1f_k-1)
		        

		(y_k-1) edge (f1_k)	
		(y_k-1) edge (f_k-1)	
		(y_k-1) edge (f_k)
		(x_k-1) edge (f1_k)
		(x_k-1) edge (f_k-1)
		(z_k-1) edge (f1_k)

		(y_k) edge (f1_k+1)	
		(y_k) edge (f_k)	
		(y_k) edge (f_k+1)
		(x_k) edge (f1_k+1)
		(x_k) edge (f_k)
		(z_k) edge (f1_k+1)		
		
		(y_k+1) edge (f_k+1)	
		(x_k+1) edge (f_k+1)
		(x_k+1) edge (f1_k+2)
		(y_k+1) edge (f1_k+2)
		(z_k+1) edge (f1_k+2)	
		
		(y_k+1) edge[dotted] (f_k+3)	
		(y_k-2)	edge[dotted] (f_k-1)

        
        
        
        
        
        ;
    
    \begin{pgfonlayer}{background}
        \node [background,
                    fit=(1x_k-1)(1y_k-1)(1z_k+1) (1f1_k+2) (1f_k+2),
					label={[align=center]left:Model \\ for  \\task $z$}] (){};                               
        \node [background,
                    fit=(x_k-1)(y_k-1)(z_k+1) (f1_k+2) (f_k+2), 
					  label={[align=center]left:Model \\ for  \\task $y$}] (){};           
                    
    \end{pgfonlayer}
\end{tikzpicture}
}}
\subfloat{
\resizebox {0.60\columnwidth} {!} {
\input{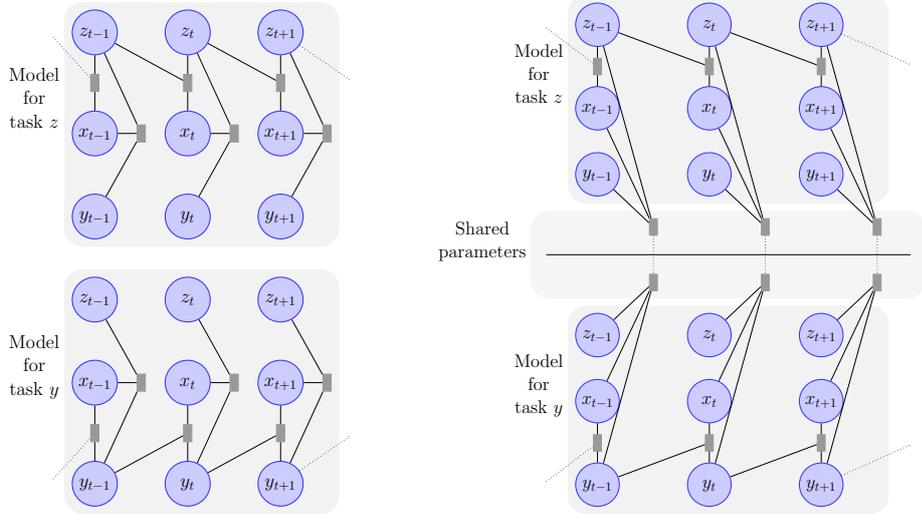}
}}
\caption{\small Graphical model representations of \textsc{unshared} (left) and \textsc{shared} (right) models. Note the common factors in the \textsc{shared} (right) model, above and below the horizontal line. These factors are defined over the same random variables and  share the parameters.}
\label{fig:shared}
\end{figure}

We model $p(\y,\z|\x)$ by considering two types of cliques (and potential function defined on those cliques). The first type of clique, similar to the one in linear chain CRF, consists of adjacent labels in {\it one} (of any) sequence $\y$, i.e., $(y_{t-1},y_t)$ along with current $x_t$ i.e. $(y_{t-1},y_t,x_t)$, and the second type of clique consists of the pair of labels $(y_t,z_t)$ along with current $x_t$ i.e. $(y_t,z_t,x_t)$. Here the first type of clique provides the independence while the second type of clique provides the benefit from other labels. As we shall see later, such a model provides {\it better discriminating power} than the models that consider all types of cliques \cite{sutton2007dynamic,marcheggiani2014hierarchical}. Given such two types of cliques and the potential functions defined over them, the conditional probability of both label sequences given the example sequence can be written as:
{\small
\begin{align}
 p^y(\y,\z|\x,\theta^y,\psi^y) =  \frac{1}{U^y(\x)} \prod_{t=1}^T \Big(  \underbrace{\Phi(y_{t-1},y_{t},x_{t}|\theta^y)}_{\text{task(y) factor}} \Big) \Big(\underbrace{\Phi(y_{t},z_{t},x_{t}|\psi^y)}_{\substack{\text{label dependency }\\\text{factor}}} \Big)
\label{eq:m1_clique}
\end{align}
}
Similar to CRF, $U^y(\x)$ is the normalization factor. 
Although \eqref{eq:m1_clique} provides the probability of both the labels, i.e., $(\y,\z)$, conditioned on observed data sequence $\x$ , there is no clique that depends on adjacent $z$ labels, i.e., $z_t, z_{t-1}$. Thus though incorporating partial information from other label $z$, the above model still focuses on the task $y$. So the above model is only defined for the task $y$ since its primary focus is label $y$ - because of the task $y$ factor.  Similarly we can define a model for task $z$:
{\small
\begin{align}
p^z(\y,\z|\x,\theta^z,\psi^z) = \frac{1}{U^z(\x)} \prod_{t=1}^T \Big( \underbrace{\Phi(z_{t-1},z_{t},x_{t}|\theta^z)}_{\text{task(z) factor}}\Big) \Big(\underbrace{\Phi(y_{t},z_{t},x_{t}|\psi^z)}_{\substack{\text{label dependency }\\\text{factor}}} \Big)
\label{eq:m2_clique}
\end{align}
}
It is to be noted in the above models that the first clique is the task-specific clique as it only considers the label from one task, while the second clique is the common clique as it takes labels from both tasks. Since the second clique is common in both models, the first clique (and label) is the model's defining clique (and label), and corresponds to the task that model is built for. Also note that in the above models each type of clique has its own parameters, i.e. task $y$ has its parameters $\theta^y$ and $\psi^y$ and the task $z$ has its own parameters $\theta^z$ and $\psi^z$. Such a model where each task has its own set of parameters, we call it \textsc{unshared} model.
A pictorial representation of this \textsc{unshared} model is shown in \figref{shared}. Observe that there are two different models, one for each task. Both models have their own factors (and parameters).

The above models can be optimized (and inferenced) using the standard machinery used in CRF since these models are exactly the same as CRF except an additional clique. 
 
%
%

 
Below we define the generalized \textsc{unshared} multilabel model, i.e., there can be any number of labels with arbitrary dependencies among them.
\begin{defn}
Let $\x$ be an observed example sequence with $\y_1,\y_2,\ldots \y_k$ its multiple label sequences. Let $\mathcal{C}_t$ be the set of cliques denoting the possible interactions among labels at time $t$ (i.e., interaction among labels $\y_1,\y_2,\ldots \y_k$), then, the \textsc{unshared} multilabel model is a set of task-specific models where each task-specific model (for task $\y_l$) is defined as:
{\small
\begin{align}
p^{y_l}(\y_1,\y_2,\ldots,\y_k|\x,\theta^{y_l},\psi^{y_l}) = \frac{1}{Z(\x)}  \Big( \prod_{t=1}^T \Phi(y_{l,{t-1}},y_{l,t},x_{t}|\theta^{y_l})\Big) \Big(\prod_{t=1}^T \prod_{c \in \mathcal{C}_t} \Phi(\y_{c},x_{t}|\psi^{y_l}) \Big) 
\label{eq:defn_unshared}
\end{align}
}
\label{defn:unshared}
\end{defn}
 
\subsection{Shared Models}
\label{sec:ap1}
Although more accurate than the existing methods (CRF and factorial CRF) (see experiments), this method does not take advantage of the multitask nature of the problem, as both models have their own separate set of parameters, and there is no learning transfer between these models. We exploit the multitask nature of the problem and facilitate learning transfer by sharing the parameters corresponding to the common clique in both models. Sharing parameters to facilitate learning transfer is a well-known practice in multitask learning \cite{Micchelli04regularizedmulti-task,jacob08,argyriouFeature06,andreas07,Argyriou07convexmultitask}. In other words, we make 
$$\psi^y = \psi^z=\psi.$$
We call this formulation \textsc{shared} model. A pictorial representation of this \textsc{shared} model is shown in \figref{shared}. We emphasize in this figure that there are two {\it separate} models, with one set of factors that is common to both models. The figure should not be confused for the graphical model for one single model. The parameters corresponding to the common factor are shared between both models, as opposed to \textsc{unshared} model where both models have their own parameters.

Now for the clarity and follow up discussion, we write the formulations \eqref{eq:m1_clique} and \eqref{eq:m2_clique} in terms of corresponding feature functions (under \textsc{shared} model):
{\small
\begin{align}
p^y(\y,\z|\x,\theta^y,\psi) = \frac{1}{U^y(\x)} \prod_{t=1}^T \exp \Big(  \sum_k \Big( \underbrace{\theta^y_k f_k(y_{t-1},y_t,x_{t})}_{\text{task(y) factor}}  + \underbrace{\psi_k f_k(y_{t},z_{t},x_t)}_{\substack{\text{label dependency }\\\text{factor}}} \Big) \Big).
\label{eq:m1}
\end{align}
}
Here $ U^y(\x) = \sum_{\y,\z} \prod_{t=1}^T \exp ( \sum_k ( \theta^y_k f_k(y_{t-1},y_t,x_{t})+ \psi_k f_k(y_{t},z_{t},x_t) ) )$.
We can write a similar model for the task $z$. In this model, first type of clique depends on the adjacent labels from task $z$ along with $x_t$ i.e. $(z_t,z_{t-1},x_t)$ while the other type of clique is similar to the model for task $y$.
{\small
\begin{align}
p^z(\y,\z|\x,\theta^z,\psi) = \frac{1}{U^z(\x)} \prod_{t=1}^T \exp \Big(  \sum_k \Big( \underbrace{\theta^z_k f_k(z_{t-1},z_t,x_{t})}_{\text{task(z) factor}}  
+ \underbrace{\psi_k f_k(y_{t},z_{t},x_t)}_{\substack{\text{label dependency }\\\text{factor}}} \Big) \Big).
\label{eq:m2}
\end{align}
}

\noindent
{\bf Remarks:} Two main points to be noted about these two \textsc{shared} models are: (a) though we have two models, one for each task, each of these models is sufficient to produce labels for both tasks, and (b) the parameters $\theta^x$, $\theta^y$ are task specific while parameters $\psi$ are common to both tasks which facilitates learning transfer among both tasks. Since each model can be used to produce labels for both tasks, these two tasks can be thought as two views, and one can use co-training with these models to build a semisupervised model. 

\begin{defn}
A \textsc{shared} multilabel model is a set of task-specific models, where each task-specific model is defined as in \eqref{eq:defn_unshared} but all parameters corresponding to the label-dependency factor are same. In other words:
$$\psi^{y_1} = \psi^{y_2}= \ldots = \psi^{y_k} = \psi.$$
\end{defn}

%
%
%
Next we construct our objective function to fit data to these models. We take four specific approaches to define objective function as described below.  


\paragraph{\bf Joint Optimization:}
We hypothesize that although each of these models are sufficient to learn the labels for both tasks independently, it will be advantageous to learn them simultaneously. Consequently, we define a joint model that is the product of both models\footnote{Note that though each of these two models gives us a probability distribution over $(\y,\z)$, product of these two models is not a probability distribution. This product is taken only to facilitate the joint learning -- a practice used in MTL \cite{Micchelli04regularizedmulti-task,argyriouFeature06,andreas07}. One can also think of maximizing this joint log likelihood as minimizing the {\it cumulative} loss of both models on the training data which is the negative of joint log likelihood.}. We maximize the likelihood of the data under this model, i.e., find the parameters by optimizing the joint log likelihood. This is equivalent to minimizing the loss on the training data.
To reduce the overfitting, we define Gaussian prior with mean $\mu=0$ and covariance matrix $\Sigma=I/\eta$ for all parameters i.e., $p(\theta^y) \propto \exp(-\frac{\eta^y}{2} \|\theta\|^2)$, $p(\theta^z) \propto \exp(-\frac{\eta^z}{2} \|\theta\|^2)$ and $p(\psi^y) \propto \exp(-\frac{\eta^o}{2} \|\theta\|^2)$. The log likelihood of the data with this modeling approach can be written as:
{\small
\begin{align}
\ell(\theta^y,\theta^z,\psi)  = \sum_{i=1}^n \log{p^y\left(\y^{(i)},\z^{(i)}|\x^{(i)},\theta^y,\psi\right)}  + \log{p^z\left(\y^{(i)},\z^{(i)}|\x^{(i)},\theta^z, \psi\right)} - \nonumber \\
\frac{\eta^y}{2} \|\theta^y\|^2 - \frac{\eta^z}{2} \|\theta^z\|^2 - \frac{\eta^o}{2} \|\psi\|^2
\label{eq:jointlik}
\end{align}
}

The derivatives of the above joint log likelihood with respect to $\theta^y$ (similar for $\theta^z$) and $\psi$ are:
{\small
\begin{align}
\frac{\partial(\ell)}{\partial{\theta^y_k}}= \sum_i \sum_t f_k(y^{(i)}_t,y^{(i)}_{t-1},x^{(i)}_t) - \sum_i \sum_t \sum_{y,y'} p^y(y,y'|\x^{(i)},\theta^y,\psi) f_k(y,y',x^{(i)}_t) - \eta^y \theta^y_k
\label{eq:der_thetay}
\end{align}
}
{\small
\begin{align}
\frac{\partial(\ell)}{\partial{\psi_k}} & = \sum_i \sum_t 2f_k(y^{(i)}_t,z^{(i)}_{t},x^{(i)}_t) - \sum_i \sum_t \sum_{y,z} p^y(y,z|\x^{(i)},\theta^y,\psi) f_k(y,z,x^{(i)}_t) \nonumber \\ 
& \qquad  \qquad - \sum_i \sum_t \sum_{y,z} p^z(y,z|\x^{(i)},\theta^z,\psi) f_k(y,z,x^{(i)}_t) - \eta^o \psi 
\label{eq:der_psi}
\end{align}
}
where, the first term is simply the feature value while the second (and third) terms are the expectations of the feature values over all possible label combinations, as is standard in log-linear models \cite{berger1996maximum,lafferty2001conditional}. Observe that computing these expectations require us to compute marginal probabilities, i.e., $p^y(y,y'|\x^{(i)},\theta^y,\psi)$, $p^y(y,z|\x^{(i)},\theta^y,\psi)$ and $p^z(y,z|\x^{(i)},\theta^z,\psi)$.

Note that the joint likelihood function $\ell(\theta^y,\theta^z,\psi)$ is convex in all its parameters i.e. $\theta^y$, $\theta^z$ and $\psi$ and hence can be optimized by a number of techniques. In our implementation, we use L-BFGS which has previous shown to outperform other techniques \cite{sha2003shallow}. For inferences, we need two kind of inferences, one for computing marginals, e.g., $p^y(y,y'|\x^{(i)},\theta^y,\psi)$ (sum-inference) and other for computing the most likely label i.e., $\arg \max_{\y,\z} p(\y,\z|\x)$ (max-inference). We use belief propagation for sun-inferences and Viterbi for max-inferences.

The above described model has some resemblance with the factorial CRF model\cite{sutton2007dynamic} (described in \secref{relatedwork}) with the important difference that the factorial CRF has one single model which is jointly optimized for all tasks and, therefore, has no explicit parameter sharing. On the other hand, we break the factorial CRF in two separate tasks and then explicitly share the parameters among both tasks. This difference is important because breaking the one model into two models increases their discriminative power (the normalization factor is also broken). Such a separate framework allows the transfer of learning through parameter sharing but at the same time, leaves enough room for independent learning. This independent learning is important as you shall see in the experiments, in some cases, \textsc{unshared} model performs better than the factorial CRF because in those cases independent learning is more important that the partial sharing as done in factorial CRF. For mathematical details on this, refer to Appendix~\ref{app:ap1}.

\paragraph{\bf Alternate Optimization:}
we propose a variation of the objective function in \eqref{eq:jointlik}. We split the joint likelihood into two parts, one for each task and optimize them alternatively. $\psi$ is still a common set of parameters among both tasks however we do not optimize the joint likelihood. Under this model, the likelihood of the data can be written as following:
{\small
\begin{align*}
\ell_y(\theta^y,\psi) = \sum_{i=1}^n \log p^y\left(\y^{(i)},\z^{(i)}|\x^{(i)},\theta^y,\psi\right) - \frac{\eta^y}{2} \|\theta^y\|^2 - \frac{\eta^o}{2} \|\psi\|^2  \\
\ell_z(\theta^z,\psi) = \sum_{i=1}^n \log p^z\left(\y^{(i)},\z^{(i)}|\x^{(i)},\theta^z,\psi\right) - \frac{\eta^z}{2} \|\theta^z\|^2 - \frac{\eta^o}{2} \|\psi\|^2
\end{align*}
}
The derivative of the above functions with respect to $\theta^y$ and $\theta^z$ are same as in \eqref{eq:der_thetay}. The derivative with respect to $\psi$ is little different from \eqref{eq:der_psi} as it will only have terms from corresponding $p^y$ or from $p^z$, depending on if we are taking derivative w.r.t to $\ell_y$ or $\ell_z$, but not from both. For our implementation, we use L-BFGS but in an alternate fashion. In alternate optimization, we take one step to optimize the parameters of first model (run only one gradient step), and then take one step of second model. The whole process is repeated for a fixed number of iterations. Note that though both likelihood functions are convex in themselves, this alternate optimization technique is not guaranteed to converge, however, in practice, it seems to converge.  Although both models give us the assignment for both labels $\y$ and $\z$, we consider the assignments from their respective models i.e. $\y$ assignment from the first model\eqref{eq:m1} and $\z$ assignment from the second model \eqref{eq:m2}.


\subsection{Variance Models}
\label{sec:variance}
The primary purpose of multitask learning framework is to be able to transfer learning among multiple tasks in a way that each model is able to model its own task, and at the same time, is also able to benefit from other tasks. We have thus far, incorporated this paradigm by having a set of parameters common among different tasks. The task specific part of each task is captured by a factor specific to that task. We extend this framework by splitting the common set of parameters (label dependency) into two parts: one task specific while other common. We hypothesize that the whole label dependency factor may not be common to both tasks, but only a part of it. As we shall see shortly that it will bring flexibility in the model, allowing one to control the amount of transfer among different tasks.

Along the lines of \cite{Micchelli04regularizedmulti-task}, we believe that the parameters corresponding to the label dependency factor lie around a common set of parameters having their own variance specific to task. With this assumption, the common set of parameters $\psi$ can be written as:
$$\psi^y = \psi^o + \nu^y$$
Now, $\psi^o$ is the part that is common to all tasks while $\nu^y$ is the task specific part. This is to indicate that there might be a component of $\psi$ that is only specific to that task when considering parameters $\psi$. The task $y$ model under this assumption can be written as following (task $z$ model will be similar):
{\small
\begin{align*}
  p^y(\y,\z|\x,\theta^y,\nu^y, \psi^o) =   \frac{1}{U^y(\x)} \prod_{t=1}^T \exp \Big( \sum_k \Big( \theta^y_k f_k(y_{t-1},y_t,x_{t})  + \nu^y_k f_k(y_{t},z_{t},x_t)  \\  +  \psi^o_k f_k(y_{t},z_{t},x_t)\Big) \Big) 
\end{align*}
}
The log likelihood under this model can be given as (with Gaussian prior on each set of parameters):
{\small
\begin{align*}
\ell_y(\theta^y,\nu^y,\psi^o) = \sum_{i=1}^n \log p^y(\y^{(i)},\z^{(i)}|\x^{(i)},\theta^y,\nu^y,\psi^o) - \frac{\eta^y}{2} \|\theta^y\|^2 - \frac{\lambda}{2} \|\nu^y\|^2 - \frac{\eta^o}{2} \|\psi^o\|^2 
\end{align*}
}
We emphasize here the importance of the weight factor $\lambda$ associated with the regularization $\|\nu\|^2$. This $\lambda$ enforces the model to share the label dependency parameters among different tasks. A high value of $\lambda/\eta^o$ means that there will be more sharing among tasks while a small value of $\lambda/\eta^o$ means that the task would be unrelated as if there is no sharing of parameters among tasks. Note that when $\lambda/\eta^o \rightarrow 0$, it will force parameters $\psi^o$ to go to zero which will results in an \textsc{unshared} model; and when $\lambda/\eta^o \rightarrow \infty$, it will force task-specific parameters $\theta^y$ and $\nu^y$ to go to zero which will result in a model that will be completely shared, i.e., same for all tasks. Therefore, one can also think of this $\lambda$ factor as an interpolating factor, interpolating between completely shared model and an \textsc{unshared} model. Under this variance model, similar to the previous models, there can be two ways to model the data likelihood: one is jointly and other alternative. 


\section{Related Work}
\label{sec:relatedwork}
Most of the work in multitask learning has focused on the standard classification or regression problems, a very few have focused on the sequence labeling problem, e.g. \cite{sutton2005composition}. In this work, authors do not use MTL setting directly. They learn each task independently  but in a cascaded manner i.e. use the output of one task as input to the other, but tests by considering all tasks simultaneously. Therefore, authors do not make use of tasks' relatedness or label dependency at the training time.
In classification and regression MTL, there are mainly two approaches, Bayesian and non-Bayesian. In both the approaches, one of the fundamental problem is defining the task relatedness and then incorporating that in the model. In MTL literature,  most of the existing methods first assume a structure that defines the task relatedness, and then incorporate this structure in the MTL framework in the form of a regularizer~\cite{andreas07,Micchelli04regularizedmulti-task,jacob08}. There are many other approaches to multitask learning such as subspace methods \cite{argyriouFeature06,andreas07,Argyriou07convexmultitask}, parameter proximity methods \cite{Micchelli04regularizedmulti-task}, and task clustering methods \cite{jacob08}. In subspace method, it is assumed that the parameters of different tasks lie in a subspace whereas in proximity method, we assume that task parameters $w_t$ for each task is close to some common task $w_0$ with some variance $v_t$. These $v_t$ and $w_0$ are learned by minimizing the {\it Euclidean} norm  which is again equivalent to working in the linear space. This idea of proximity method is later generalized through manifold regularization \cite{agarwal2010learning} and clustering \cite{jacob08}. 

For the sake of completeness we give a brief description of Factorial CRF, which will also be our primary baseline. Factorial CRF ~\cite{sutton2007dynamic} model is an extension of linear-chain CRFs that repeat structures and parameter over sequences. If we denote by $\Phi_c(y_{c,t}, x_t)$ the repetition of clique c at time step t, then a factorial CRF defines the probability of a label sequence $\y$ given the input $\x$ as: 
{\small
$$p(\y|\x) = \frac{\prod_t \Phi_c ({\bf y}_{c,t},{\bf x}_t)} {{\bf Z(x)}}$$
}
Factorial CRF can be generalized to model connection between multiple label sequences, i.e. $\y_l$ for $l=\{0,1,..,L\}$ for the same input sequence $\x$. Sutton, et al.,~\cite{sutton2007dynamic} defines the $p(\y|\x)$ distribution as below:
{\small
$$p(\y|\x) = \frac{1}{{\bf Z(x)}} \left(\prod_{t=1}^{T-1}\prod_{l=1}^{L} \Phi_l(y_{l,t},y_{l,t+1},{\bf x,t})\right)  \left(\prod_{t=1}^{T}\prod_{l=1}^{L-1} \Psi_l(y_{l,t},y_{l+1,t},{\bf x,t})\right)$$  
}
where $\Phi_l()$ is the task specific factor that models the dependency between the consecutive tokens in a label sequence whereas $\Psi_l()$ is the label dependency factor that models the dependency among labels of the same token. Although factorial CRFs model the dependencies among multiple labels in a sequence, it considers the whole learning problem as one single task. Enforcing this one task structure on the problem constrains the problem, leaving little room for independent learning from multiple labeling tasks. On the other hand, in our method, we break the problem into multiple tasks, allowing room for flexibility for independent learning from both label dependency factor $\Psi_l$ and task specific factor $\Phi_l$, and at the same time benefiting from each other through explicit parameter sharing.

\section{Experiments}
\label{sec:exp}
In this section, we describe the datasets, our experimental methodology, and report results.

\subsection{Dataset} We evaluate and report our results on two datasets. First dataset corresponds to a noun phrase chunking and POS tagging tasks, and comes from a CoNLL 2000 shared-task \footnote{Publicly available at \cite{tjong2000introduction}\url{http://mallet.cs.umass.edu/grmm/data}}.
We take a smaller set of the original data set primarily because MTL only makes sense when single task learning (STL) is not sufficient (i.e. it is difficult). This difficulty of STL can be attributed to two main reasons-- one, there are not enough labeled examples, and second, the problem itself is a difficult problem despite being enough labeled examples. The CoNLL dataset violates both of these conditions, i.e., there are enough labeled examples, and these labeled examples give a very good accuracy i.e., in the range of $99\%$. So in order to make the MTL applicable here, we increase the difficulty of the problem by reducing the size of labeled data. The smaller dataset consists of total 350 sentences containing 8785 individual tokens as examples. We split the data into 150 train and 200 test examples. In this dataset, two tasks correspond to the NP chunking and part-of-speech (POS ) tagging. The idea is to get performance improvement by learning from these two tasks simultaneously. This dataset is also used in the baseline method by Sutton et al.,~\cite{sutton2007dynamic}. For the sake of completeness, we also ran our experiments on full dataset, and all methods performed between $98\%$ and $99\%$. 

The second dataset comes from an electronic conversation medium over social media (twitter). The example set is borrowed from real conversations (chat) between customers and customer care agents for a particular telecommunication carrier. Two specific tasks are designed in this case where the chat sentences are labeled for (1) nature of dialogue between customer and agent (namely {\it Dialogue Act}), and (2) nature of the state of the issue being discussed by customer and agent (namely {\it Issue Status}). We employed 3 annotators for labeling each sentence present in the conversations. Each conversation is treated as a sequence example akin to a sentence in the first dataset. For first task, sentences are annotated from 12 labels: Complaint, Apology, Answer, Receipt, General Compliment, Directed Compliment, Request, Greeting, Thank, Announcement, Solved, and Other. For second task, sentences are annotated with 4 labels: Open Issue, Issue Resolved, Change Medium of Communication, and Issue Closed. We take 291 annotated conversations with a total of 3072 sentences with 10.6 sentences per conversation. We append frequent bigrams, emoticons, punctuation and standard word features such as capitalization etc.    

\subsection{Models Comparisons}
We use following models for comparisons. Among these models, one is baseline, other models are ours, with different variations.
\begin{itemize}
\item {\bf Factorial CRF\cite{sutton2007dynamic}:} We use this as our primary baseline. 
\item {\bf Unshared model:} Both tasks have their own separate parameters (See \defnref{unshared}). 
\item {\bf JOSP:} (Jointly Optimized Shared Parameters) This is the shared model where parameters are learned by optimizing the joint likelihood.
\item {\bf AOSP:} (Alternatively Optimized Shared Parameters) This is the shared model where parameters are learned alternatively as described in  \secref{ap1}.
\item {\bf JOVM:} (Jointly Optimized Variance Model) Variance model as defined in \secref{variance} but parameters are learned  by optimizing the joint likelihood.
\item {\bf AOVM:} (Alternatively Optimized Variance Model) Variance model as defined in \secref{variance} but parameters are learned alternatively.
\end{itemize}

\subsection{Results}
We use accuracy as our metric of evaluation. Here we define accuracy as fraction of correctly labeled tokens in sequences present in the test set. It is important to note that we report the accuracy from their respective models i.e., each model gives labels for all tasks but we take the labels from the model that is specific to that task (as described in \secref{ap1}). 

The results for the two datasets are presented in \tabref{mallet} and \ref{tab:conv}. The subtables in each table corresponds to individual tasks. We vary the training size and report the results each subtable. All reported results are averaged over 10 random runs. From these results we draw multiple conclusions: (1) In general, learning tasks together in MTL setting ---either directly or using variance method--- helps. All results show significant improvement over factorial CRF. This improvement is higher when there are fewer labeled examples. (2) Though in some cases, MTL (Shared model and Variance model) helps over factorial CRF but  learning them independently (\textsc{Unshared} model) helps even more. e.g. Issue Status task. This establishes the fact that not all tasks improve from MTL. In fact, it shows that in multiple tasks, one task can benefit from other tasks while another cannot.

From the accuracy figures, it can be inferred that the Task 1 is harder than Task 2 for both datasets. The results reported show that the accuracy improvements are greater for Task 1 compared to Task 2. For difficult tasks, results show that learning both tasks independently  (\textsc{Unshared} model) hurts. Learning them together through explicit parameter sharing gives significant improvement over \textsc{Unshared} or factorial CRF. This observation along with the observation that MTL improvement is higher when there are fewer labeled examples, provide evidence in support of the hypothesis about the applicability of MTL, i.e.,  MTL is applicable when the underlying problem is difficult, either inherently or because of the scarcity of labeled examples. The results are not as clear for Task 2, but still, in these tasks, results indicate that one should use MTL -- either learn all tasks together through {\it explicit} parameter sharing (Shared model or Variance model) or not share anything at all (\textsc{Unshared model}). Partial sharing (one task structure) as in fatorial CRF gives inferior results. 

\begin{figure}[htb]
\centering
\subfloat{\includegraphics[width=.35\textwidth]{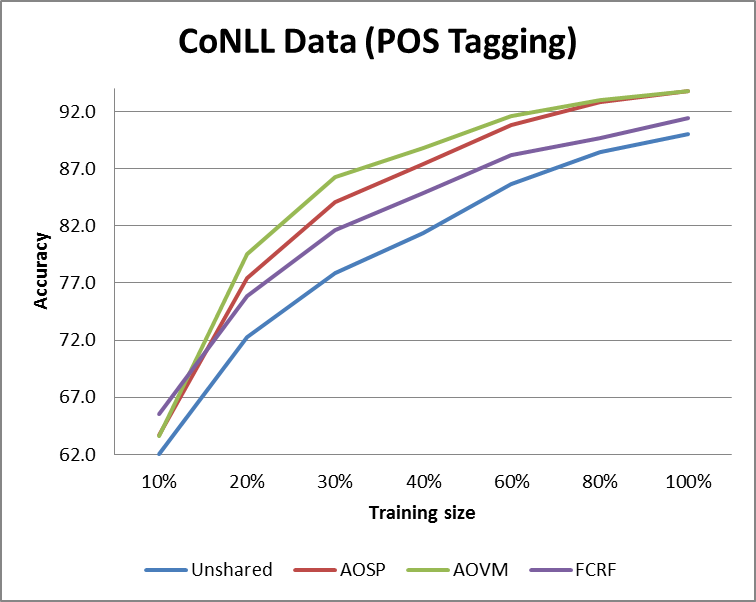}}
\subfloat{\includegraphics[width=.35\textwidth]{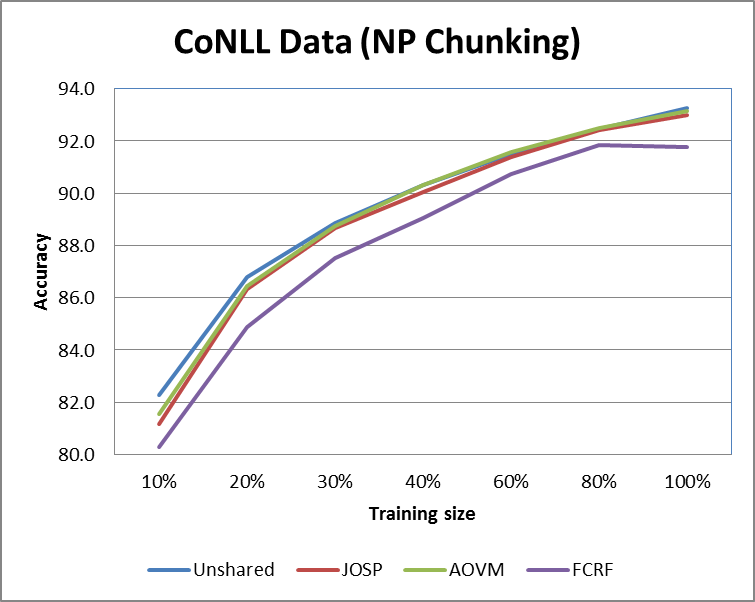}}
\caption{Variation with training size for CoNLL data}
\label{fig:var_train_conll}
\end{figure}

\begin{figure}[htb]
\centering
\subfloat{\includegraphics[width=.35\textwidth]{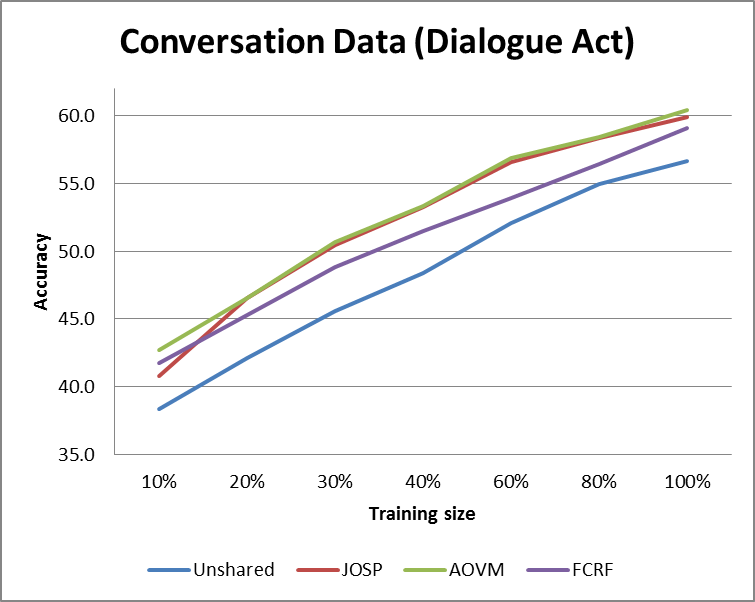}}
\subfloat{\includegraphics[width=.35\textwidth]{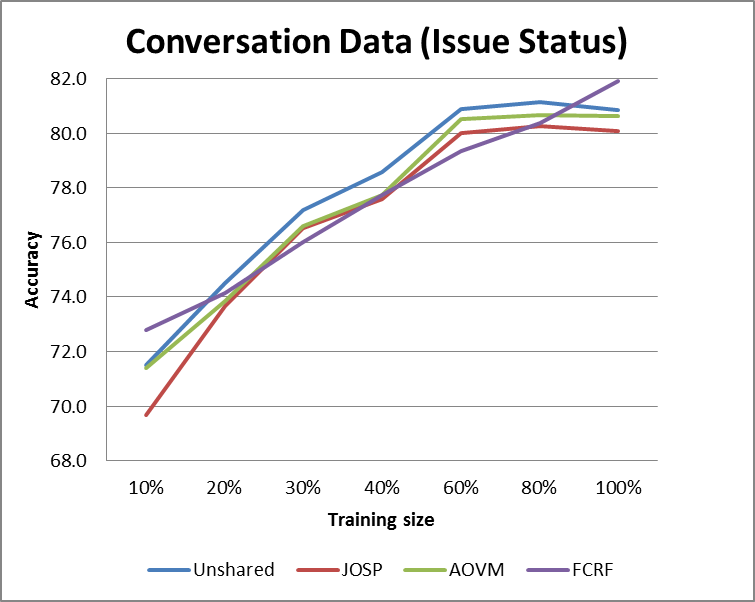}}
\caption{Variation with training size for conversation data}
\label{fig:var_train_conv}
\end{figure}

We also varied the training size and recorded the accuracies. The results are plotted in \figref{var_train_conll} and \figref{var_train_conv}, for CoNLL and conversation data respectively. These figure also support our earlier hypothesis. From these figures, we see that when tasks are difficult (Task 1), MTL models (variance and shared) perform better, but when tasks are rather easy, \textsc{Unshared} model performs better. 


\begin{table}[ht]
\tiny
\centering
\caption{Dataset:CoNLL 2000 NP Chunking}
\subfloat[\small Task 1 : POS Tagging]{%
    \hspace{.5cm}%
	\begin{tabular}{|c|ccccc|c|}
	\hline 
	\%Train&    \multicolumn{5}{|c|}{MTL}   & DCRF  \\ 
	\cline{2-6}
   & JOVM  & AOVM & JOSP &  AOSP & Unshared &  \\ 
	\hline 
	$(30\%)$ &$86.0$ & ${\bf 86.3}$& $83.7$ & $84.1$ & $77.9$  & $81.6$  \\ 
	\hline
	$(60\%)$ & $91.5$& ${\bf 91.6}$ & $90.7$ & $90.8$ & $85.7$  & $88.2$  \\ 
	\hline
	\end{tabular} 
}
\subfloat[\small Task 2 : NP Chunking]{%
    \hspace{.5cm}%
	\begin{tabular}{|c|ccccc|c|}
	\hline 
	\%Train&    \multicolumn{5}{|c|}{MTL}   & DCRF  \\ 
	\cline{2-6}
   & JOVM  & AOVM & JOSP &  AOSP & Unshared &  \\ 
   \hline 
	$(30\%)$ & ${\bf 89.0}$ &$88.8$ & $88.5$ & $88.7$ & $ 88.8$  & $87.5$  \\ 
	\hline
	$(60\%)$ & $91.5$ &${\bf 91.6}$ & $91.3$ & $91.4$ & $91.5$  &  $90.7$ \\ 
	\hline
	\end{tabular} 
 }
\label{tab:mallet} 
\end{table}

\vspace{-0.7in}

\begin{table}[ht]
\tiny
\centering
\caption{Dataset:Conversation}
\subfloat[\small Task 1 : Dialogue Act]{%
    \hspace{.5cm}%
	\begin{tabular}{|c|ccccc|c|}
	\hline 
	\%Train&    \multicolumn{5}{|c|}{MTL}   & DCRF  \\ 
	\cline{2-6}
   & JOVM  & AOVM & JOSP &  AOSP & Unshared &  \\ 
   \hline 
	$(30\%)$ & ${\bf 51.4}$ &$50.7$ &$45.3$  & $50.5$ & $45.6$  & $48.9$  \\ 
	\hline
	$(60\%)$ & $56.7$ &${\bf 56.9}$ &$55.7$ & $56.6$ & $52.1$  & $53.9$  \\ 
	\hline
	\end{tabular} 
}
\subfloat[\small Task 2 : Issue Status]{%
    \hspace{.5cm}%
	\begin{tabular}{|c|ccccc|c|}
	\hline 
	\%Train&    \multicolumn{5}{|c|}{MTL}   & DCRF  \\ 
	\cline{2-6}
   & JOVM  & AOVM & JOSP &  AOSP & Unshared &  \\ 
   \hline 
	$(30\%)$ & ${\bf 77.2}$ & $76.6$ & $74.4$ & $76.5$ & ${\bf 77.2}$  & $76.0$  \\ 
	\hline
	$(60\%)$ & $80.3$ & $80.5$ & $80.8$ & $80.0$ &  ${\bf 80.9}$ & $79.4$  \\ 
	\hline
	\end{tabular} 
 }
\label{tab:conv} 
\end{table}

\vspace{-2cm}

\section{Conclusion}
In this paper, we have presented a novel method for learning from multiple sequence labeling tasks. Unlike the previous methods, our method models each task as one single model, but still transfer the learning from other tasks through parameters sharing. We have shown through various experiments on two datasets that our method consistently outperforms the one of the best methods for such tasks, especially in cases when tasks are relatively harder and there are few labeled examples. One additional advantage of our method is that unlike most methods in MTL in which each model only learns on its own labels (and hence outputs its own labels only), the proposed method learns using all labels which makes this approach extensible for semi-supervised setting through co-training. 

\appendix
\section{Factorial CRF and Shared Joint Model}
\label{app:ap1}
{
For reference, we write below the factorial CRF model:
{\small
\begin{align*}
 p(\y,\z|\x,\theta^y,\theta^z, \psi) &= \frac{1}{U(\x)} \prod_{t=1}^T \exp\Bigg(\sum_k \Big( \underbrace{\theta^y_k f_k(x_{t},y_{t-1},y_t)}_{\text{task(y) factor}} + \underbrace{\theta^z_k f_k(x_{t},z_{t-1},z_t)}_{\text{task(z) factor}} \nonumber \\
& \qquad \qquad + \underbrace{\psi f_k(x_t,y_{t},z_{t})}_{\text{label dependency factor}}\Big) \Bigg)
\end{align*}
}

It might look like that the above factorial CRF model is similar to the product of the two models $p_y$ and $p_z$ which it is not. This is because of the normalization factor in individual model. The product of two models (\eqref{eq:m1} and \eqref{eq:m2})can be written as:
{\small
\begin{align*}
& q(\y,\z|\x,\theta^y,\theta^z \psi)  = p^y(\y,\z|\x,\theta^y,\psi) \; p^z(\y,\z|\x,\theta^z,\psi) \\
& \qquad  = \frac{1}{U'(\x)} \prod_{t=1}^T \exp \Bigg( \sum_k \theta^y_k f_k(x_{t},y_{t-1},y_t) + \theta^z_k f_k(x_{t},z_{t-1},z_t) 
+ 2\psi_k f_k(x_t, y_{t},z_{t}) \Bigg)
\end{align*}
}
where 
{\small
\begin{align*}
U'(\x) & = U^y(\x) \; U^z(\x) \\
& =\Bigg( \sum_{\y,\z} \prod_{t=1}^T \exp \Bigg(  \sum_k \Big( \theta^y_k f_k(x_{t},y_{t-1},y_t) 
 + \psi_k f_k(x_t, y_{t},z_{t}) \Big) \Bigg) \Bigg)  \Bigg( \sum_{\y,\z} \prod_{t=1}^T \exp \Bigg(  \\
& \qquad \qquad + \sum_k \Big( \theta^z_k f_k(x_{t},z_{t-1},z_t)+ \psi_k f_k(x_t, y_{t},z_{t}) \Big) \Bigg) \Bigg)\\
& \ne U(\x).
\end{align*}
}
Note that in the above product of the two models, the numerator is very similar to the factorial CRF, (they are the same except that the common factor $\psi$ is counted twice) but the denominator is completely different. The denominator in factorial CRF {\it cannot} be written as the product of the denominator of two models, i.e., $U'(\x) \ne U(\x)$. This breaking of numerator is important because it allows the model to break into multiple tasks hence allowing for independent learning, at the same time facilitating transfer learning through parameter sharing. 
}

{\small
\bibliography{mtl_seq_ref}
\bibliographystyle{styFiles/splncs03}
}
\end{document}